%% file: aist.tex
\documentclass[runningheads]{llncs}
\usepackage{graphicx}
\usepackage{easyeqn}
\usepackage{subcaption}
\usepackage{amsfonts}
\def\T{\mathrm{T}}

\def\mv{\mathbf{m}}

\def\xv{\mathbf{x}}

\def\Ev{\mathbf{E}}
\def\Mv{\mathbf{M}}
\def\Wv{\mathbf{W}}

\begin{document}
\title{Dropout Strikes Back: Improved Uncertainty Estimation via Diversity Sampling}
\titlerunning{Dropout Strikes Back}
%
\author{Kirill Fedyanin\inst{1} \and
Evgenii Tsymbalov\inst{1,2}\and
Maxim Panov\inst{1}}

%
\authorrunning{K. Fedyanin, E. Tsymbalov, M. Panov}
%
\institute{Skolkovo Institute of Science and Technology (Skoltech), Moscow, Russia \\
\email{\{k.fedyanin, m.panov\}@skoltech.ru}\\
\and
Yandex, Moscow, Russia\\
\email{etsymba@yandex-team.ru}}
\maketitle              
\begin{abstract}
  Uncertainty estimation for machine learning models is of high importance in many scenarios such as constructing the confidence intervals for model predictions and detection of out-of-distribution or adversarially generated points. In this work, we show that modifying the sampling distributions for dropout layers in neural networks improves the quality of uncertainty estimation. Our main idea consists of two main steps: computing data-driven correlations between neurons and generating samples, which include maximally diverse neurons. In a series of experiments on simulated and real-world data, we demonstrate that the diversification via \textit{determinantal point processes}-based sampling achieves state-of-the-art results in uncertainty estimation for regression and classification tasks. An important feature of our approach is that it does not require any modification to the models or training procedures, allowing straightforward application to any deep learning model with dropout layers.

\end{abstract}

\section{Introduction}
\label{sec:intro}
\input{intro}

\section{Methods}
\label{sec:main}
\input{methods}

\section{Experiments}
\label{sec:experiments}
\input{experiments}

\section{Related Work}
\label{sec:related_work}
\input{related_work}

\section{Conclusions}
\label{sec:conclusions}
\input{conclusions}


\subsubsection*{Acknowledgements}
   The research was carried out at Skoltech and supported by the Russian Science Foundation (project no. 21-11-00373). The authors want to thank Nikita Mokrov for useful discussions. M.P. and K.F. acknowledge the use of ``Zhores'' supercomputer~\cite{Zacharov2019} for obtaining the part of results presented in this paper.

\bibliographystyle{splncs04}
\bibliography{uncertainty}

\end{document}


\maketitle

\appendix

\section{Regression Experiments}
\label{sec:regression_suppl}
\input{regression_suppl}

\section{Classification Experiments}
\label{sec:classification_suppl}
\input{classification_suppl}

\bibliographystyle{apalike}
\bibliography{uncertainty}

%% file: intro.tex

Uncertainty estimation (UE) recently became a very active
area of research in deep learning. Neural networks
usually are treated as black boxes, and in general, they
are prone to overconfidence~\cite{guo2017,gal2016}. Uncertainty estimation methods aim to help overcome this drawback by identifying potentially erroneous predictions. This can be especially important for error-critical applications like medical diagnostics~\cite{begoli2019need} or autonomous car driving~\cite{feng2018towards}. 
Another important application for uncertainty estimation is active learning~\cite{Settles2012}.
The majority of sampling criteria in active learning are based on estimates of uncertainty, which makes it important to obtain high-quality uncertainty estimates.

There are several main approaches for uncertainty estimation for deep neural networks. Bayesian neural networks (BNN) and variational inference in particular represent a natural way for uncertainty estimation due to availability of well-defined posteriors, but they can be prohibitively slow for large-scale applications. The usage of dropout at the inference stage was shown to be good and efficient approximation to BNNs~\cite{Gal2015,gal2016}. The ensembles of independently trained models~\cite{lakshminarayanan2017simple} have state-of-the-art performance in many tasks requiring uncertainty estimation~\cite{snoek2019can}. Recently, forcing models in ensembles to be more diverse was shown to improve results even further~\cite{jain2019maximizing}. The drawback of ensembles is that we need to train and use multiple models that require additional resources, i.e. more memory to store models and more computing power for training.

In this work, we aim to develop a new approach for dropout-based uncertainty estimation.
Usually there are many highly correlated neurons in neural networks, which results in a slow convergence of estimates based on the standard uniform sampling in dropout layers.
We propose to estimate correlations between neurons based on the data and sample the most diverse neurons in order to improve the convergence of the estimates and, as a result, the quality of uncertainty estimates. 
As a particular realization of the general idea, we suggest sampling dropout masks using the machinery of determinantal point processes (DPP)~\cite{kulesza2012determinantal} which are known to give diverse samples.

We summarize the main contributions of the paper as follows:
\begin{itemize}
  \item We propose two DPP-based sampling methods for neural networks with dropout. Our approach requires to train only a single model and adds only small overhead on the inference stage compared to plain MC dropout.
  
  \item We compare different dropout-based approaches for uncertainty estimation in an extensive series of experiments for real-world regression and classification datasets. The results show superior performance of proposed DPP-based approaches.

  \item Importantly, the proposed methods show high quality of uncertainty estimation even for very small number of stochastic passes through the network, thus opening the possibility to significantly speed up the inference stage.

\end{itemize}

The rest of the paper is organized as follows. Section~\ref{sec:main} introduces the proposed method for DPP-based sampling from neural networks with dropout. In Section~\ref{sec:experiments}, we show the efficiency of the proposed approach in the problem of uncertainty estimation. Section~\ref{sec:related_work} gives an overview of the related work on uncertainty estimation for neural networks. Section~\ref{sec:conclusions} concludes the study and highlights some directions for future work.

%% file: methods.tex

\subsection{Neural Networks with Dropout as Implicit Ensembles}
  We start by considering a standard fully connected layer in a neural network
  \begin{EQA}[c]
  \label{eq:linear_layer}
    \textstyle{S_{i}^{h} = \sum_{j = 1}^{N_{h - 1}} w_{ij}^{h} O_{j}^{h-1}, ~~ i = 1, \dots, N_{h},}
  \end{EQA}
  where \(O_{i}^{h} = \sigma\bigl(S_{i}^{h}\bigr)\) is an output of the \(h\)-th layer of the neural network given by a non-linear transformation \(\sigma(\cdot)\) of the corresponding pre-activation \(S_{i}^{h}\).
  %
  An application of dropout to neurons results in the following formula for the pre-activations:
  \begin{EQA}[c]
  \label{eq:dropout_linear}
    \textstyle{S_{i}^{h} = \sum_{j = 1}^{N_{h - 1}} \frac{1}{1 - p} m_{j}^{h} w_{ij}^{h} O_{j}^{h-1}, ~~ i = 1, \dots, N_{h},}
  \end{EQA}
  where \(m_{j}^{h}\) are Bernoulli random variables with a probability of \(0\) equal to \(p\). The outputs \(O_{i}^{h}\) of the \(h\)-th layer remain to be computed by \(O_{i}^{h} = \sigma\bigl(S_{i}^{h}\bigr)\).
  Note that if an input variable of neural network is denoted by \(\xv\), then then output of every layer is a function of \(\xv\), i.e., \(O_{i}^{h} = O_{i}^{h}(\xv)\).

  Let us denote the vector of dropout weights \(m_{j}^{h}\) for the \(h\)-th layer by \(\mv_h = (m_{1}^{h}, \dots, m_{N_h}^{h})^{\T}\) and the full set of dropout weights by \(\Mv = (\mv_1, \dots, \mv_K)\). Thus, any neural network \(\hat{f}(\xv)\) with dropout layers essentially has two sets of parameters: the full set of learnable weights \(\Wv\) and the set of dropout weights \(\Mv\): 
  \begin{EQA}[c]
    \hat{f}(\xv) = \hat{f}(\xv \mid \Wv, \Mv).
  \end{EQA}

  Let us have a neural network with dropout, which was trained on some dataset giving weight estimates \(\hat{\Wv}\). Then 
  dropout weights \(\Mv\) can be considered as free parameters and require selection at the time of inference 
  \begin{EQA}[c]
    \hat{f}(\xv \mid \Mv) = \hat{f}(\xv \mid \hat{\Wv}, \Mv).
  \end{EQA}
  The originally proposed~\cite{hinton2012} and currently the standard choice is to take \(\hat{\Mv} = (1 - p) \cdot \Ev\), where \(\Ev\) is the matrix of all ones of the corresponding shape. Such an approach gives the fixed function \(\hat{f}(\xv \mid \hat{\Mv})\), which is known to give reasonably good performance in practice. The main intuition behind such choice is the replacement of the stochastic pre-activations \(S_{i}^{h}\) given by~\eqref{eq:dropout_linear} with their expectations, which are exactly equal to~\eqref{eq:linear_layer}.

  Recently, it was proposed to consider dropout as a variational approximation in a specially chosen Bayesian model, see~\cite{Gal2015}.
  %
  Within this approach, one can sample 
  \(T\) 
  i.i.d. realizations \(\Mv_1, \ldots, \Mv_T \sim Bernoulli(1 - p)\) and compute approximate posterior mean and variance 
  \begin{EQA}[c]
    \bar{f}_T(\xv) = \frac{1}{T} \sum_{i = 1}^{T} \hat{f}(\xv \mid \Mv_i), \quad
    \bar{\sigma}_T^2(\xv) = \frac{1}{T} \sum_{i = 1}^{T} \bigl(\hat{f}(\xv \mid \Mv_i) - \bar{f}_T(\xv)\bigr)^2.
  \label{eq:mean_variance}
  \end{EQA}
  
  The approximate posterior variance \(\bar{\sigma}_T^2(\xv)\) is a natural choice for the uncertainty estimate and was successfully used in the variety of applications such as out-of-distribution detection~\cite{vyas2018out} and active learning~\cite{gal2017deep}.

  In this paper, we suggest a different approach, namely we treat \(\hat{f}(\xv \mid \Mv)\) as an ensemble of models indexed by dropout masks \(\Mv\). Such a view allows us to decouple inference from training and pose an intuitive question: what set of masks \(\Mv_1, \ldots, \Mv_T\) should one choose in order to obtain the best uncertainty estimate \(\bar{\sigma}_T^2(\xv)\)?

  Importantly, here we do not limit the selection of masks to be samples from standard dropout distribution, which, in principle, should allow us to obtain better estimates. However, the design of mask selection procedure is a non-trivial problem, which we discuss below in detail.

  \begin{remark}
    The standard approach in the literature is to consider an ensemble of models trained on different subsets of the data set or just from different random initializations giving the set of parameter estimates \(\hat{\Wv}_1, \ldots, \hat{\Wv}_T\) and corresponding approximations \(\hat{f}(\xv \mid \hat{\Wv}_i, \hat{\Mv}), ~ i = 1, \ldots, T\). Similarly, one can compute the variance \(\bar{\sigma}_T^2(\xv)\), which was shown to be a reasonable uncertainty estimate in practice~\cite{smith2018less,Beluch2018}. The main drawback of this approach is the need to train and store \(T\) different models, which might be very costly both in terms of computation and storage needed.
  \end{remark}

\subsection{Data-driven Mask Generation Under General Sampling Distributions}

\begin{figure}[t!]
    \begin{subfigure}[t]{0.5\textwidth}
      \centering
      \includegraphics[width=0.75\textwidth]{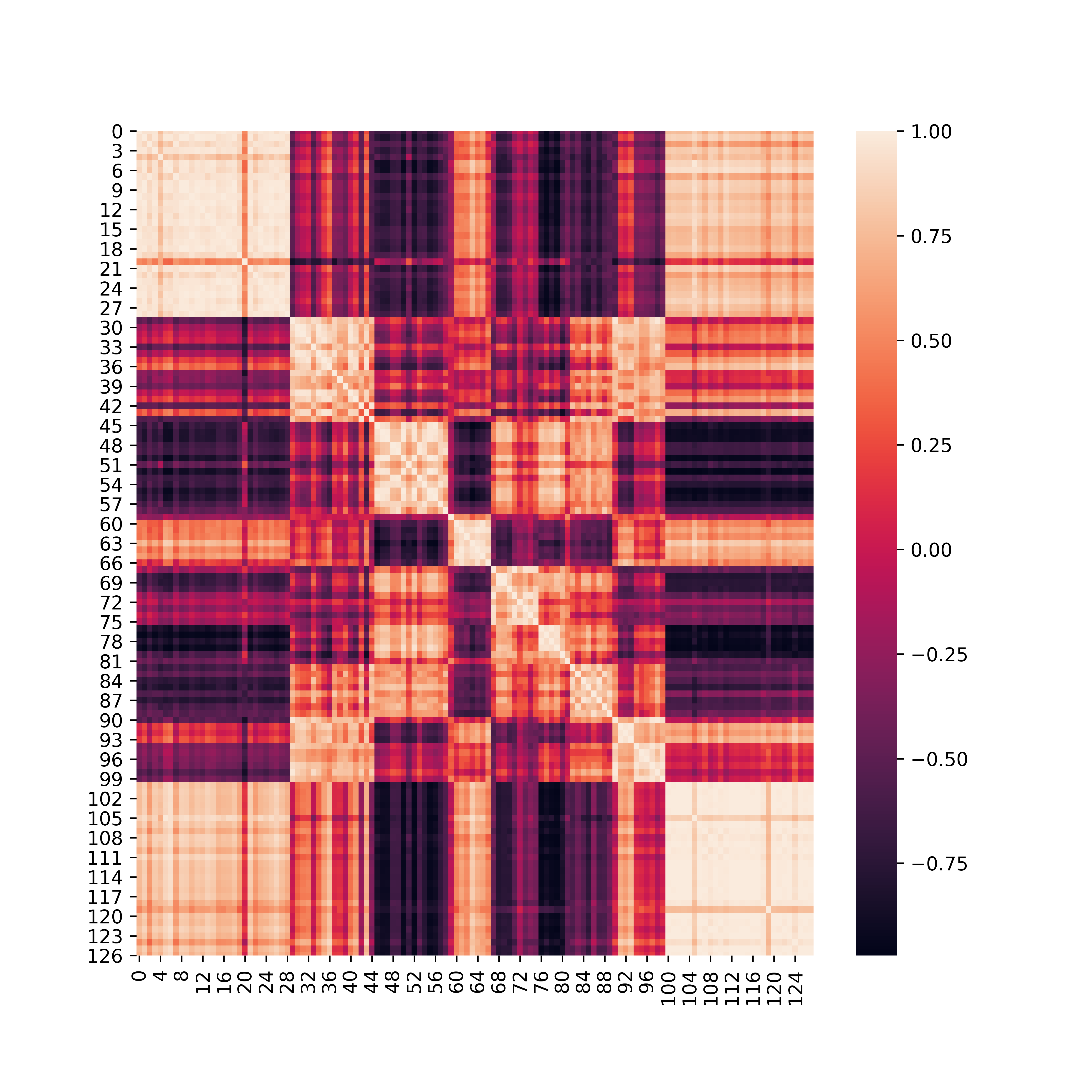}
      \caption{Correlation matrix.}
    \label{fig:naval-corr}
    \end{subfigure}
    \begin{subfigure}[t]{0.5\textwidth}
      \includegraphics[width=\textwidth]{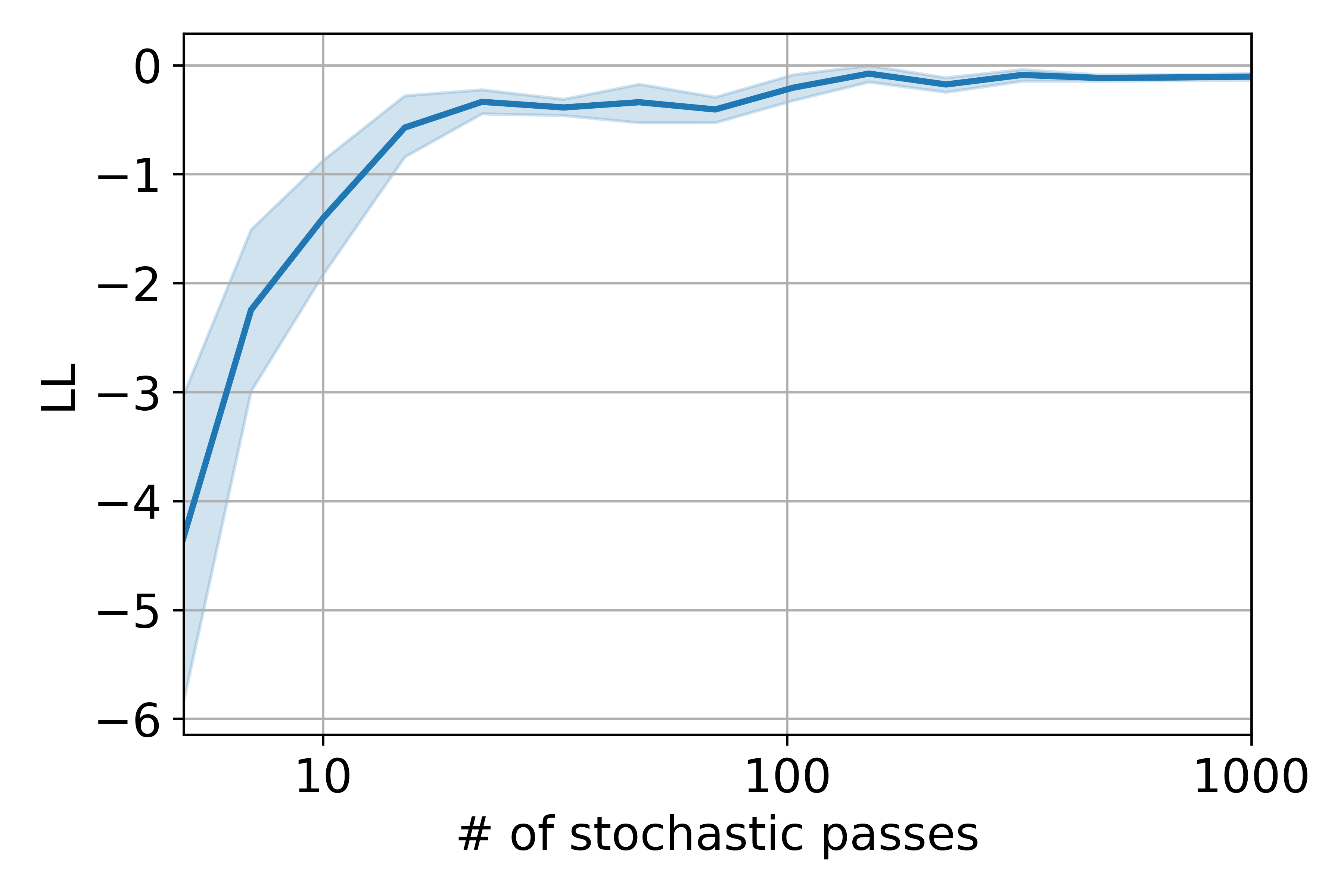}
      \caption{Log-likelihood for MC dropout as a function of \(T\).}
    \label{fig:ll_convergence}
    \end{subfigure}
    \caption{\textbf{(a)} Correlation matrix \(C\) between the outputs of the neurons in a hidden layer of the NN trained on the \textit{naval propulsion} dataset.
    \textbf{(b)} For the same dataset log-likelihood computed via MC dropout increases with increase of the number of stochastic passes \(T\). More than 100 samples are needed to reach convergence.}
  \end{figure}

  In practice, many neurons in the network are highly correlated. For example, consider a correlation matrix of neurons in a hidden layer of a fully-connected neural network, trained on the regression dataset (see Figure~\ref{fig:naval-corr}). The correlation matrix was computed on the test set and clearly shows groups of highly correlated neurons. Sampling masks for this layer uniformly at random might result in a high variance of pre-activations~\eqref{eq:dropout_linear}. As a result, the estimates for the whole network may require a significant number of samples (stochastic passes through the NN) \(T\) to converge. We illustrate this behaviour on Figure~\ref{fig:ll_convergence}, where 
  several hundreds of simple MC dropout estimates are required for the convergence of the log-likelihood values. It is clearly seen that a larger number of samples improves the values of log-likelihood, yet may impose computational cost too large to be used in real-world applications.
   However, one may expect that the knowledge about the correlations between neurons can help to sample more diverse neurons and improve the estimates.

  In what follows, we consider the probabilistic generation of masks \(\mv_h\) from some distribution \(P^{(h)}\) with possibly non-i.i.d. distributions of components. Similarly to the case of dropout, we suggest using an unbiased estimate of the layer-wise mean. Our main motivation is to approximately preserve the average performance of the trained network. The construction of the unbiased estimator is non-trivial and is given by celebrated Horvitz-Thompson (HT) estimator~\cite{Horvitz1952}:
  \begin{EQA}[c]
  \label{eq:ht_layer}
    \textstyle{S_{i}^{h} = \sum_{j = 1}^{N_{h - 1}} \frac{1}{\pi_j^h} m_{j}^{h} w_{ij}^{h} O_{j}^{h-1}, ~~ i = 1, \dots, N_{h},}
  \end{EQA}
  where \(\pi_j^h\) is the marginal probability of value \(1\) for the random variable \(m_{j}^{h}\).

\subsection{Diversity Sampling Approaches}
\label{sec:diversity}
  Let us consider \(h\)-th hidden layer of the neural network with dropout. Assume that we have access to the correlations 
  \begin{EQA}[c]
    C_{ij}^{(h)} = \mathrm{corr}_{\xv} \bigl\{O_{i}^{h}(\xv), O_{j}^{h}(\xv)\bigr\}, ~ i, j = 1, \dots, N_h.
  \end{EQA}
  In practice, we compute an empirical correlation based on some set of points, which represents the data distribution well enough. As a result, we obtain the correlation matrix \(C^{(h)} \in \mathbb{R}^{N_h \times N_h}\) between the neurons of the \(h\)-th hidden layer.  Below we discuss several approaches to sampling neurons in a way that the correlation between sampled neurons is as small as possible. We note that instead of the correlation matrix \(C^{(h)}\) one may consider the covariance matrix \(K^{(h)}\) in any of the approaches described below. The properties of the methods significantly depend on the choice of the matrix, and we will perform the empirical evaluation of the methods based on each of them in the experiments.

\subsubsection{Leverage Score Sampling}
\label{sec:decorr}
  A basic approach for non-uniform sampling of rows and columns in kernel matrices is the so-called \textit{leverage score sampling}~\cite{alaoui2015fast}. 
  In this approach, the neurons are sampled independently with different probabilities \(\pi_j^{h}\):
  \begin{EQA}[c]
    \pi_j^{h} \sim \ell_{\lambda}^{(h)}(j) = \Bigl[C^{(h)} \bigl(C^{(h)} + \lambda I\bigr)^{-1}\Bigr]_{jj}, ~ j = 1, \dots, N_h,
  \end{EQA}
  \noindent
  where the quantities \(\ell_{\lambda}^{(h)}(j)\) are called \textit{leverage scores}.
  %
  This approach makes neurons from large and highly correlated clusters to be sampled less frequently. In Section~\ref{sec:experiments}, we show that leverage score sampling indeed allows obtaining better uncertainty estimates for out-of-distribution data in regression tasks compared to MC dropout. However, its performance for in-domain data is even inferior to uniform sampling. In the next section, we propose a more complex approach, which allows to significantly improve the quality of uncertainty estimation.

\subsubsection{Sampling with Determinantal Point Processes}
\label{sec:dpp}
  Determinantal Point Processes (DPPs)~\cite{kulesza2012determinantal} are specific probability distributions over configurations of points that encode diversity through a kernel function. They were introduced in~\cite{macchi1975coincidence} for the needs of statistical physics and were used for a number of ML applications, see~\cite{kulesza2012determinantal} for an overview. DPP can be seen as a probabilistic MaxVol algorithm~\cite{goreinov2010find} of finding a maximal-volume submatrix.

  We use correlation matrix \(C^{(h)}\) as the likelihood kernel for DPP. Then, given a set \(S\) of selected points for a mask distribution \(\mv_h \sim DPP\bigl(C^{(h)}\bigr)\), we obtain
  \begin{EQA}[c]
    \mathbb{P}[\mv_h = S] = \frac{\det \bigl[C^{(h)}_S\bigr]}{\det \bigl[C^{(h)} + I\bigr]}, ~~ h = 1, \dots, K,
  \end{EQA}
  where \(C_S^{(h)} = \Bigl[C_{ij}^{(h)}, ~ i, j \in S\Bigr]\), i.e., a square submatrix of \(C^{(h)}\) obtained by keeping only rows and columns indexed by \(S\). 

  To better understand the DPP, let us come back to the correlation matrix depicted in Figure~\ref{fig:naval-corr}. The probability for DPP to take highly correlated neurons into the sample \(S\) is low as, in this case, the corresponding determinant \(\det C^{(h)}_S\) will have a small value. Thus, DPP tends to sample neurons from different clusters, increasing an overall diversity.

  From computational point of view, DPP-sampling requires \(O(N_{h}^3)\) operations for generating each sample. It is quite expensive but completely viable even for modern large networks which usually have up to 1024 neurons in fully-connected layers. Importantly, masks can be precomputed once, and then the same masks are used on the inference stage for every test sample with no additional overhead. Also, computations in last fully-connected layers with dropout usually require only few percents of the total computational budget in ImageNet-size networks. Therefore, a computational overhead caused by the DPP-sampling does not have a significant impact on the inference time.

\subsubsection{k-DPP}
\label{sec:kdpp}
  The k-DPP~\cite{kulesza2012determinantal} is a variation of the DPP, conditioned to produce samples of fixed size \(|S| = k\). With the cost of introducing an additional parameter, it allows us to tune the sampling procedure as the choice of \(k\) apparently has a significant influence on the result. In this work, we use for the \(h\)-th layer \(k^{(h)} = (1 - p) N_h\), so that the number of neurons in the sample is equal to the mean number of neurons in the sample of MC-Dropout. In the case of k-DPP, the computation of the marginal probabilities \(\pi_j^{h}\) for HT-estimator~\eqref{eq:ht_layer} is non-trivial and requires the separate optimization procedure, see the details in~\cite{Amblard2018}.

\subsection{Diversification for Uncertainty Estimation in Classification}
  For regression, the variance of prediction is a standard uncertainty measure. However, uncertainty estimation for classification is, in some sense, more challenging than for regression as there is no obvious candidate for uncertainty measure.

  Let us define the average probability for the class prediction by ensemble members \(\bar{p}_T(y = c \mid \xv) = \frac{1}{T} \sum_{i = 1}^{T} p(y = c \mid \xv, \Mv_{i})\). The standard uncertainty measure usually considered in the literature is 
  \begin{EQA}[c]
    s(\xv) = 1 - \max\limits_c ~ \bar{p}_T(y = c \mid \xv),
  \end{EQA}
  which is based solely on the mean probabilities predicted by the ensemble. While providing good results in practice~\cite{snoek2019can,ashukha2019pitfalls} it doesn't use the information about the variation of predictions between ensemble members.  

  In our work, we consider \textit{BALD}~\cite{houlsby2011bayesian} uncertainty measure and combine it with different sampling schemes considered above. BALD is equal to the mutual information between outputs and model parameters:
  \begin{EQA}[c]
    \textstyle{I(\xv) = H(\xv) - \frac{1}{T} \sum_{c = 1}^{C} \sum_{i=1}^{T} -p(y = c \mid \xv, \Mv_{i}) \log\bigl(p(y = c \mid \xv, \Mv_{i})\bigr),}
  \end{EQA}
  where \(H(\xv) = -\sum_{c = 1}^{C} \bar{p}_T(y = c \mid \xv) \log\bigl(\bar{p}_T(y = c \mid \xv)\bigr)\) is an entropy of the ensemble mean. Importantly, BALD values are directly linked with the diversity of the ensemble members, and therefore are well suited for combination with our approach. 

%% file: experiments.tex

\subsection{Uncertainty Estimation for Regression}

\subsubsection{Models and Metrics}
  For the experiments, we consider MC dropout as a baseline and all the proposed UE methods discussed in the Section~\ref{sec:diversity}: leverage score sampling, DPP and k-DPP. We present the results for leverage score sampling and DPP based on correlation matrix and k-DPP based on covariance matrix as such a choices give consistently better results compared to an alternative. 
  For leverage score sampling we deliberately choose \(\lambda = 1\) to make it working with de-facto the same matrix as DPP-based methods. 
  All the regression models were trained with RMSE as a loss function.
  We used feed-forward NNs with 3 hidden layers (128-128-64 neurons) and leaky ReLU activation function~\cite{leaky_relu}. For DPP-based methods, we use the DPPy implementation provided in~\cite{GaBaVa18}.

  We should note that we do not compare with fully Bayesian approaches as we are focusing on the solutions applicable to the standard dropout-based models without changing model architecture and training procedure.  Following~\cite{hernandez2015probabilistic,jain2019maximizing}, we use log-likelihood of Gaussian distribution with mean and variance computed by different methods as a quality measure.
  
  On top of single models, we also consider a straightforward ensemble approach with NNs trained exactly the same way as single models but from different random initializations. Our experiments show that uncertainty estimates based on ensembling of networks without sampling in individual networks doesn't work for well for the considered regression datasets.

\subsubsection{Experiments on Regression Datasets}
  Similarly to~\cite{jain2019maximizing}, we run a series of experiments on various regression datasets, see Table~\ref{tab:datasets} for the full list of datasets. We start with in-domain uncertainty estimation: for each dataset, random 50\% of points were used for training and other 50\% for testing. The log-likelihood values are averaged over testing set. Multiple experiments are done via 5 random train-test splits, 2-fold cross-validation and 5 runs of the training procedures for every model (resulting in 50 average log-likelihood values contributing to each boxplot). Uncertainty estimates were computed for different number of stochastic passes \(T=10, 30\) and \(100\) for every model.

  \begin{table}[t!]
    \centering
    \caption{Summary of the UCI datasets used in experiments, see~\cite{uci}.}
    \label{tab:datasets}
    \scalebox{0.9}{
      \begin{tabular}{|c|c|c|c|c|c|c|c|c|}
        \hline
        \textbf{Dataset name} & naval propulsion & concrete & Boston housing &  kin8nm & ccpp & red wine \\
        \hline
        \hline
        \textbf{Samples} & 11934 & 1030 & 506 & 8192 & 9568 & 1599 \\
        \hline
        \textbf{Features} & 16 & 8 & 13 & 8 & 4 & 11 \\ 
        \hline
      \end{tabular}
    }
  \end{table}

\begin{figure}[t!]
    \centering
    \includegraphics[width=1.0\textwidth]{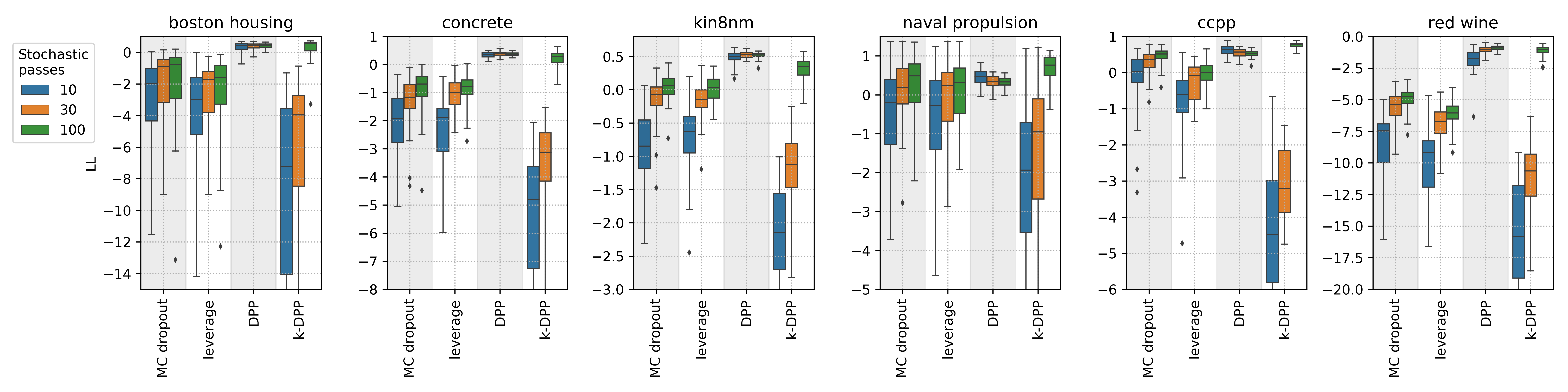} 
    \caption{Log-likelihood metric across various UCI datasets for NN UE models with different number of stochastic passes \(T = 10, 30, 100\). DPP and k-DPP give better results compared to other methods with DPP working well already for \(T = 10\) and consistently showing lower variance.}
  \label{fig:ll-single}
  \end{figure}

  We show the resulting distributions of log-likelihood values for each dataset on Figure~\ref{fig:ll-single}. We observe that either DPP or k-DPP always show the best results. Most importantly, DPP works very well already for small number of stochastic passes \(T = 10\) and consistently has low variance which is extremely important for practical usage. 

  We also performed an experiment with out-of-distribution (OOD) data. To generate OOD data we pick a random feature and split the data into the train set and OOD set by the median value on this feature. The experiments were run for 5 different splits. For OOD data good uncertainty estimates should have on average higher values compared to in-domain data. Table~\ref{tab:ood_concrete} provides for \textit{concrete} dataset the percentages of OOD points with UE values higher than $\alpha$ percentile of UE distribution for training data ($\alpha = 80\%, 90\%, 95\%$). The resulting numbers should be considered with a significant grain of salt due to their high variance but still DPP and k-DPP show the best results based on average values. 

  \begin{table}[t!]
    \caption{Percentages of OOD points with UE values higher than specified percentile of UE distribution for training data for \textit{concrete} dataset. DPP and k-DPP show the best results based on average values (top-2 average values are put in \textbf{bold}). For all the methods \(T = 100\).}
    \centering
    \vspace{5px}
    \begin{tabular}{|c|c|c|c|c|}
      \hline
      percentile & MC dropout & leverage & DPP & k-DPP \\
      \hline
      80 & 55.0$\pm$27.6 & 61.3$\pm$27.7 & \textbf{70.4}$\pm$26.0 & \textbf{71.9}$\pm$28.0 \\
      \hline
      90 & 46.0$\pm$30.7 & 52.9$\pm$30.8 & \textbf{59.6}$\pm$30.1 & \textbf{60.8}$\pm$33.7 \\
      \hline
      95 & 40.6$\pm$32.1 & 46.5$\pm$33.1 & \textbf{52.1}$\pm$32.9 & \textbf{51.8}$\pm$36.3 \\
      \hline
    \end{tabular}
  \label{tab:ood_concrete}
  \end{table}

\subsection{Uncertainty Estimation for Classification}
\label{sec:experiments_class}

\subsubsection{Data, Models and Metrics}
  In this section, we aim to show the applicability of the proposed methods to the classification tasks.
  We take \emph{BALD}~\cite{houlsby2011bayesian} as an uncertainty estimate. We consider three datasets: MNIST, which is a toy dataset of handwritten digits~\cite{lecun1998mnist}, CIFAR-10, which is a 10-class image dataset with simple objects~\cite{krizhevsky2009learning}, and ImageNet~\cite{Deng2009}, the large scale image classification dataset. Importantly, for MNIST we use only 500 train samples, otherwise the models would have too good accuracy and uncertainty estimation for in-domain data would not be relevant. For CIFAR-10 we use 50'000 samples for training and 10'000 for testing. For the MNIST dataset, we use a simple convolutional neural network with two convolutional layers, max-pooling and two fully connected layers. For the CIFAR-10 we use a more powerful network with 6 convolutional layers and batch normalization. Finally, for ImageNet we use the pre-trained ResNet-18 neural network~\cite{He2016} from PyTorch~\cite{Paszke2019}. Dropout with rate \(p = 0.5\) is used before the last fully-connected layer in all the cases. \(T = 100\) stochastic passes were made for every model. The experiments are repeated three times with different seeds for the models. 

\subsubsection{Experimental Results}
\begin{figure*}[t!]
    \centering
    \begin{subfigure}[t]{0.3\textwidth}
      \centering
      \includegraphics[height=1.3in]{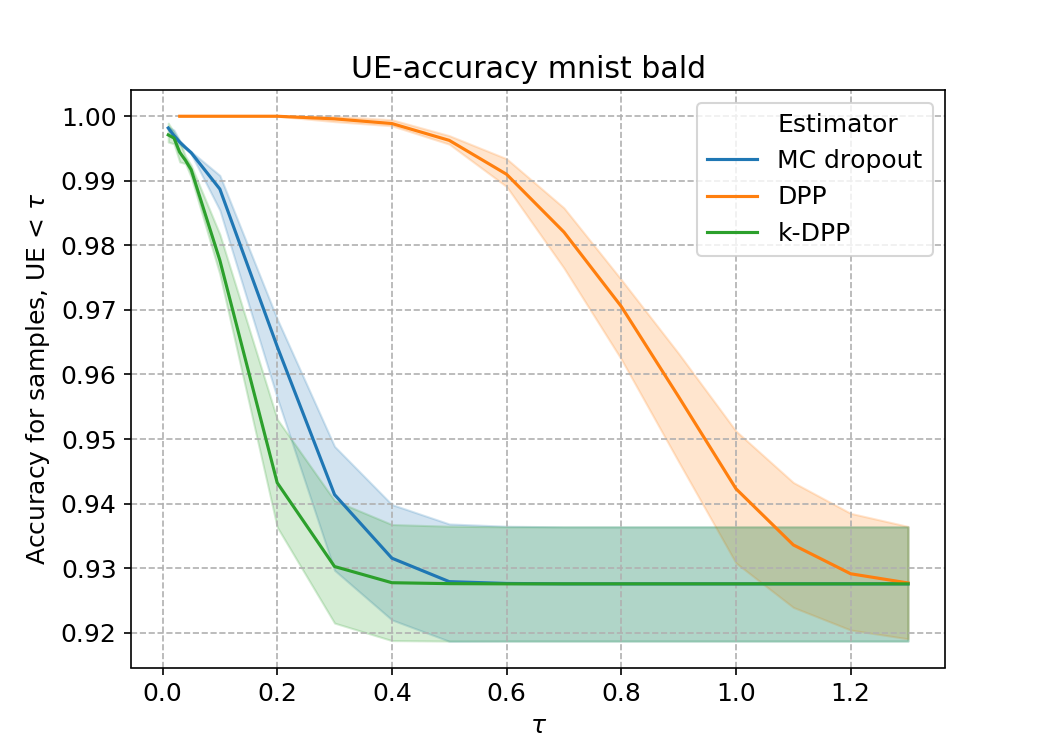}
      \caption{MNIST}
    \end{subfigure}%
    ~ 
    \begin{subfigure}[t]{0.3\textwidth}
      \centering
      \includegraphics[height=1.3in]{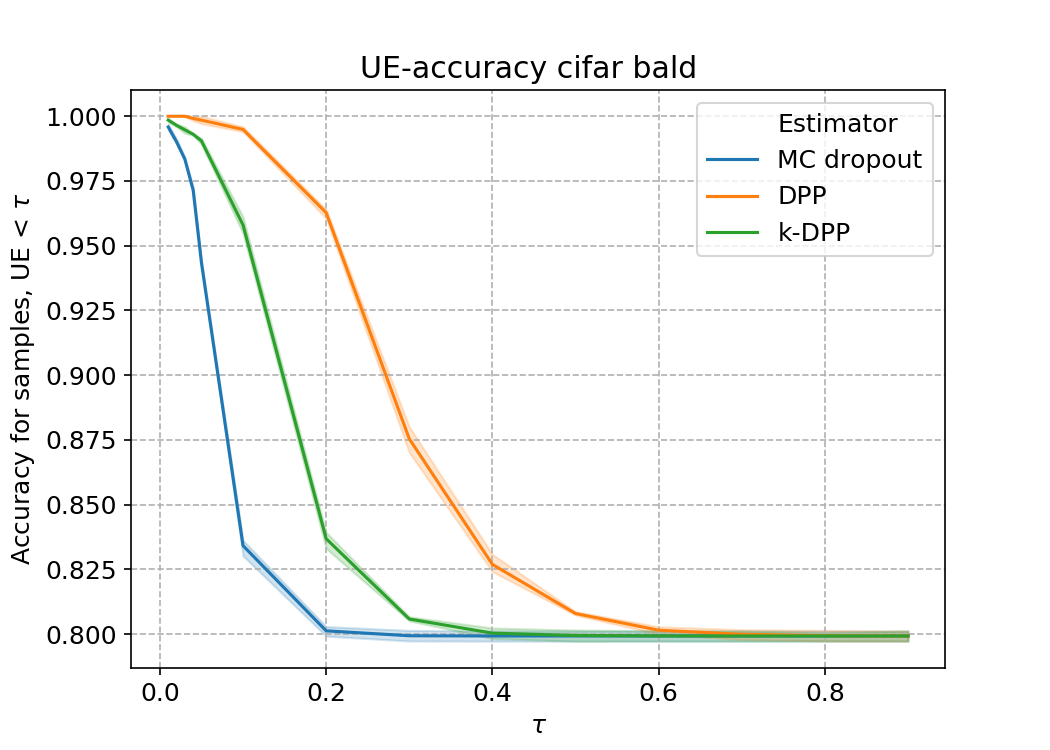}
      \caption{CIFAR}
    \end{subfigure} 
    ~
    \centering
    \begin{subfigure}[t]{0.3\textwidth}
      \centering
      \includegraphics[height=1.3in]{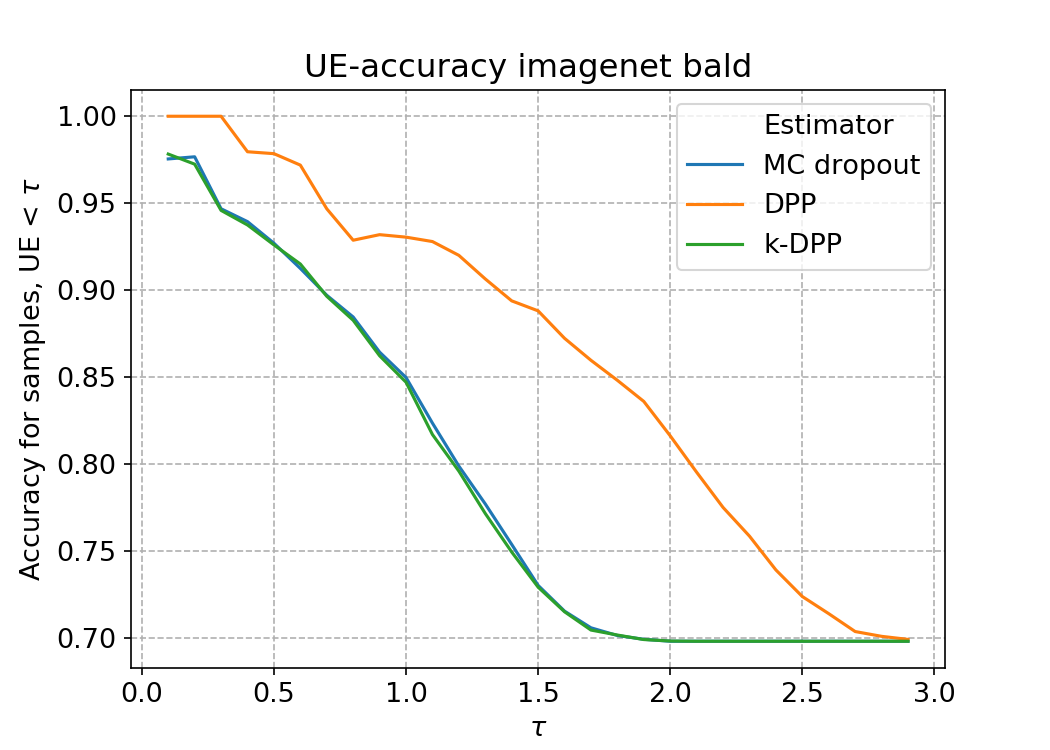}
      \caption{ImageNet}
    \end{subfigure}%

    \caption{UE-accuracy curve (the higher curve -- the better). We select the samples with low uncertainty to assure that the accuracy is higher for them.}
  \label{fig:classification}
  \end{figure*}

  For in-domain uncertainty estimation the results are presented via UE-accuracy curve, see Figure~\ref{fig:classification}. It assumes that samples with lower uncertainty will be classified with a higher average accuracy. It can be clearly seen that DPP significantly outperforms all the competitors on every dataset. We should emphasize that the superiority of DPP is especially strong for ImageNet, where the usage of DPP required only 2\% computational overhead compared to MC dropout according to our experiments.
  
\begin{figure*}[t!]
    \centering
    \begin{subfigure}[t]{0.5\textwidth}
      \centering
      \includegraphics[height=1.8in]{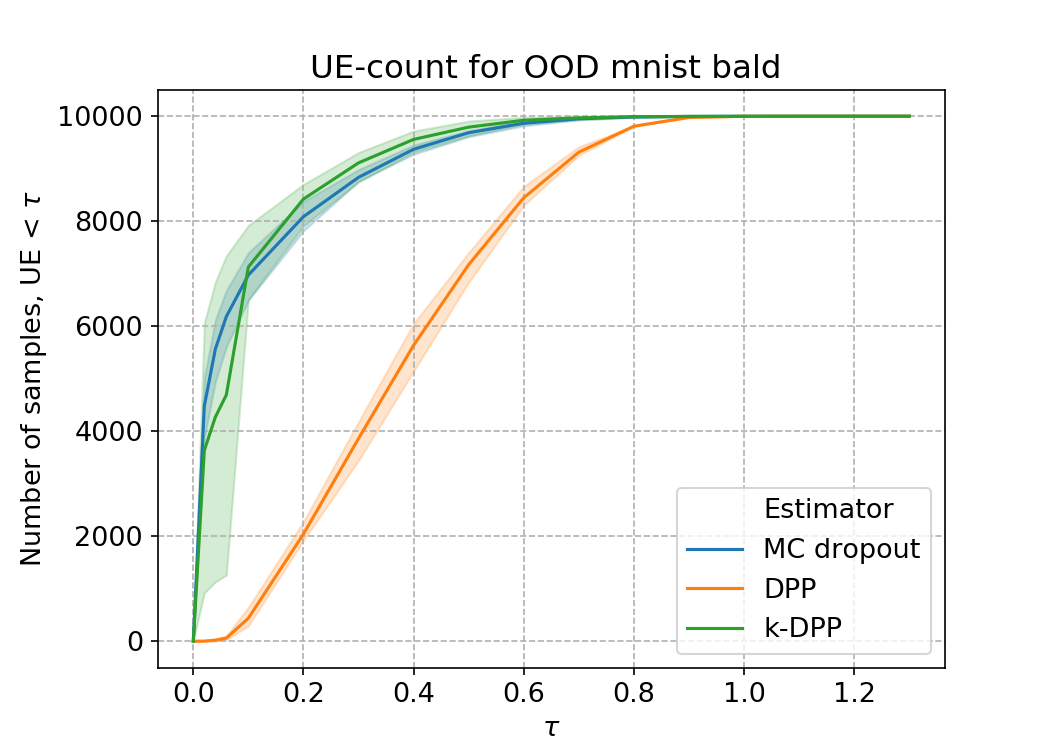}
      \caption{MNIST}
    \end{subfigure}%
    ~ 
    \begin{subfigure}[t]{0.5\textwidth}
      \centering
      \includegraphics[height=1.8in]{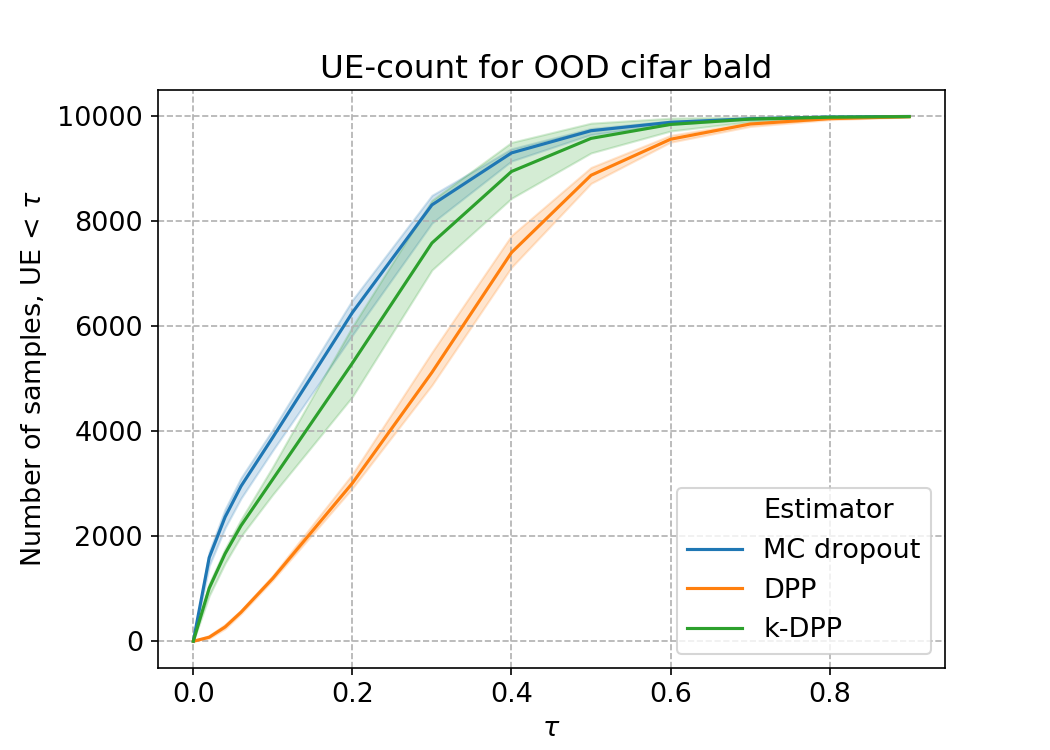}
      \caption{CIFAR}
    \end{subfigure}
    \caption{Count-vs-uncertainty curve for out-of-distribution data (the lower curve -- the better).}
  \label{fig:classification-ood}
  \end{figure*}

  We also consider detection of out-of-distribution samples which is one of the important problems for the uncertainty estimation. As OOD samples we use fashion-MNIST~\cite{xiao2017/online} and SVHN images~\cite{netzer2011reading} for MNIST and CIFAR-10 correspondingly. We use count-vs-uncertainty curve and expect there should be few points with the low uncertainty for good uncertainty estimation methods. The results are presented in Figure~\ref{fig:classification-ood}. We see that DPP-based approach allows to detect the OOD samples better for the both considered datasets.

%% file: related_work.tex

Dropout~\cite{hinton2012,srivastava2014} has emerged in recent years as a technique to prevent the overfitting in deep and overparametrized neural networks. Over the years, it obtained theoretical explanations as an averaged ensembling technique~\cite{srivastava2014}, a Bernoulli realization of the corresponding Bayesian neural network~\cite{Gal2015} and a latent variable model~\cite{maeda2014}. It was shown in~\cite{gal2016,Nalisnick2019} that using dropout at the prediction stage (i.e., stochastic forward passes of the test samples through the network, also referred to as \textit{MC dropout}) leads to unbiased Monte-Carlo estimates of the mean and variance for the corresponding Bayesian neural network trained using variational inference. These uncertainty estimates were shown to be efficient in different scenarios~\cite{gal2016,Tsymbalov2018}.

Training an ensemble of models and uncertainty estimation by their disagreement is another common approach~\cite{lakshminarayanan2017simple}. It is shown that with few models in an ensemble, you can get robust and useful calibrated results~\cite{Beluch2018}, outperforming MC dropout in active learning and error detection. The main disadvantage of ensembles is the necessity to train multiple model instances. However, it was addressed in recent works~\cite{Maddox2019,Garipov2018,Izmailov2019} which consider different strategies for speeding up ensemble construction. Recently, it was shown that improving diversity of ensemble members improves the quality of the resulting uncertainty estimates~\cite{jain2019maximizing}. We also mention recent works which thoroughly investigate in-domain~\cite{ashukha2019pitfalls} and out-of-domain~\cite{snoek2019can} uncertainty estimation in classification for the case of maximum probability uncertainty estimate.

%% file: conclusions.tex

We have proposed a new approach that strengthens the dropout-based uncertainty estimation for neural networks. Instead of randomly sampling the dropout masks on the inference stage, we sample special sets of diverse neurons via determinantal point processes that utilize the information about the correlations of neurons in the inner layers. Numerical experiments on a wide range of regression and classification tasks show that uncertainty estimates based our approach outperform the MC dropout and other baselines with a significant margin.
A combination of dropout-based inference with ensembling of several models allows to further improve the quality of the proposed uncertainty estimates and achieve state-of-the-art performance. 
From the practical perspective, our method is simple to implement as it does not require any modifications to the neural network architecture and the training process. Importantly, the proposed uncertainty estimates have high quality even for a small number of stochastic passes through the network making the inference stage even faster in practice.

We expect that the proposed methods of dropout mask sampling may also be used on the training stage, leading to more robust and efficient models. Another compelling direction of further research is approximate DPP sampling, which may increase the sampling speed of the proposed approaches, making them more production-friendly, as in \cite{shelmanov-etal-2021-certain}.

The code reproducing the experiments is available at Github\footnote{https://github.com/stat-ml/dpp-dropout-uncertainty/}.

%% file: version_2020/regression_suppl.tex

\subsection{Details on Model Training}

  
  Neural networks were trained for 10`000 epochs maximum, with checking the error on the validation set every 100 iterations: early stopping triggers if the error did not decrease for a five consecutive checks (patience = 5). Batch size equals to 500, dropout applied after the hidden layers only (except for the last layer) with rate equal to 0.5, except for the experiment C (see below). MSE was used as a loss function, and optimization was performed with the standard settings of PyTorch Adadelta optimizer. For the details on activation function and architecture please refer to Table~\ref{tab:exp-ref}, which summarizes the architectures we used (referred as Experiments A, B, C and D).

  Ensembles of models were trained separately on the same data from different random weight initializations.

\subsection{Experiments with Ensembles}
\begin{figure}[ht!] 
    \centering
    \includegraphics[scale=.35]{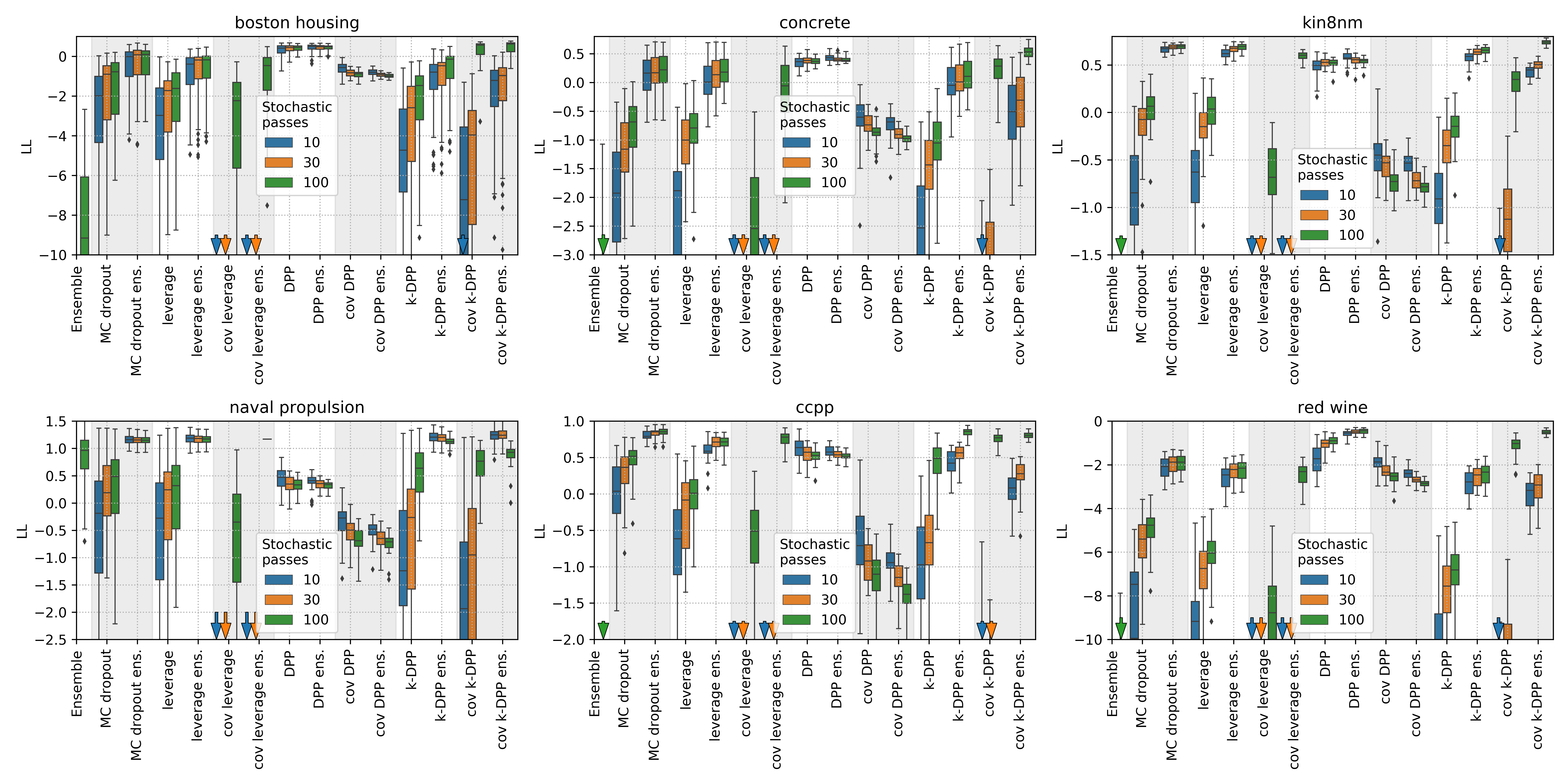}
    \caption{Log-likelihood across various UCI datasets for single models and ensembles of NN UE models. Arrows on the bottom indicate box plots being below the bottom boundary. DPP constantly shows good results in the single model scenario, being not far from ensemble-based methods.}
  \label{fig:ll-ens}
  \end{figure}

  We start with the Experiment A (the same as in the main text), consider ensembles of 5 models and combine them with the different inference methods for individual models. We visualize the log-likelihood metric for each dataset, see Figure~\ref{fig:ll-ens}. There is no single method which gives the best results uniformly over the considered datasets, yet DPP-based methods show superior performance more often than other approaches. Also it is clearly seen that pure ensembling without sampling in individual models is usually inferior even to the plain MC dropout, while combination of ensembling with sampling consistently improves the quality of uncertainty estimation.

\subsection{OOD Results for Other Datasets}

 Tables~\ref{tab:ood_boston} and~\ref{tab:ood_red_wine} provide percentages of OOD points with UE values higher than $\alpha$ percentile of UE distribution for training data ($\alpha = 80\%, 90\%, 95\%$) for two other datasets: \textit{Boston housing} and \textit{red wine}. As for the similar experiment in main text, the resulting numbers should be considered with a significant grain of salt due to their high variance but still DPP and k-DPP show the best results based on average values.
  
  \begin{table}[h!]
  \centering
    \caption{Percentages of OOD points with UE values higher than specified percentile of UE distribution for training data for \textit{Boston housing} dataset. DPP and k-DPP show the best results based on average values (top-2 average values are put in \textbf{bold}). For all the methods, \(T = 100\).}
    \begin{tabular}{|c|c|c|c|c|c|c|} \hline
    percentile & MC dropout & leverage & DPP & k-DPP \\
      \hline
80         & 49.6$\pm$26.9  & 68.1$\pm$23.7 & \textbf{69.2}$\pm$29.3 & \textbf{83.1}$\pm$23.2 \\ \hline
90         & 36.9$\pm$27.9  & 53.6$\pm$26.9 & \textbf{59.6}$\pm$31.5 & \textbf{63.7}$\pm$29.4 \\ \hline
95         & 28.2$\pm$26.7  & 40.7$\pm$30.0 & \textbf{53.5}$\pm$32.5 & \textbf{50.9}$\pm$36.0 \\
      \hline
    \end{tabular}\label{tab:ood_boston}
  \end{table}
    
  \begin{table}[h!]
  \centering
    \caption{Percentages of OOD points with UE values higher than specified percentile of UE distribution for training data for \textit{red wine} dataset. DPP and k-DPP show the best results based on average values (top-2 average values are put in \textbf{bold}). For all the methods, \(T = 100\).}
    \begin{tabular}{|c|c|c|c|c|c|c|}
      \hline
      percentile & MC dropout & leverage & DPP & k-DPP \\ \hline
80 & 50.4$\pm$26.9 & 53.1$\pm$22.3 & \textbf{73.5}$\pm$23.9 & \textbf{60.9}$\pm$26.5 \\ \hline
90 & 36.6$\pm$28.0 & 39.1$\pm$23.5 & \textbf{61.8}$\pm$29.2 & \textbf{45.0}$\pm$31.1 \\ \hline
95 & 27.3$\pm$27.7 & 30.1$\pm$22.6 & \textbf{51.0}$\pm$31.9 & \textbf{34.7}$\pm$34.2 \\ \hline
    \end{tabular}\label{tab:ood_red_wine}
  \end{table}

\subsection{Experiments with Different NN architectures}
  In order to test the robustness of the obtained results with respect to changes of NN architecture and other parameters, we settled out three more experiments with variations in:
  \begin{itemize}
    \item \textbf{architecture}. Different problems may require different number and different size of fully-connected layers of NNs to be used in order to being able fit the train data well and do not overfit.
    
    \item \textbf{activation function}. It was shown~\cite{hein2019relu} that the choice of activation function may significantly influence the quality of uncertainty estimates. We consider ReLU and CELU activation functions.
    
    \item \textbf{dropout rate}. While usally in the literature the dropout rate $p=0.5$ is often considered, for real problems lesser values of $p$ are often used in order to speed up the convergence. For example, , the dropout rate is proposed to be chosen in a cross-validation round~\cite{gal2016} together with other hyperparameters, such as regularization weight, learning rate, etc. In this work, we simply try other dropout value to check the robustness with respect to the change of this parameter.
  \end{itemize}

  The variations in the settings are provided in Table~\ref{tab:exp-ref}, with the results of Experiment A provided in the main text.
  \begin{table}[h]
    \centering
    \caption{Settings for UCI experiments.}
    \label{tab:exp-ref}
    \scalebox{1.}{
      \begin{tabular}{|c|c|c|c|c|c|c|}
        \hline
        \textbf{Index} & \textbf{Architecture} & \textbf{Activation} & \textbf{$p$} \\ \hline
        A (main text) & 128-128-64 & leaky ReLU & 0.5 \\ \hline
        B & 32-32-16 & leaky ReLU & 0.5 \\ \hline
        C & 128-128-256 & CELU & 0.2 \\ \hline
        D & 256-256-512 & CELU & 0.5 \\ \hline
      \end{tabular}
    }
  \end{table}
  
\begin{figure}[t!] 
    \centering
    \includegraphics[scale=.5]{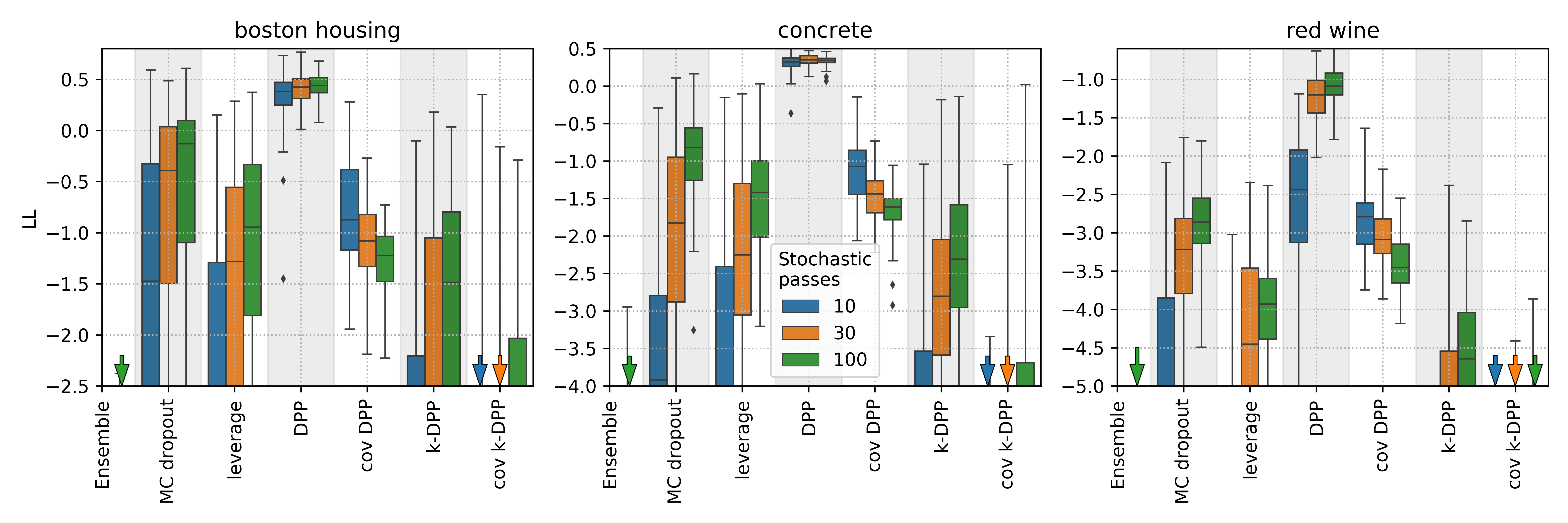}
    \caption{Log-likelihood across various UCI datasets for single models of NN UE for the small-NN experiment B, see Table \ref{tab:exp-ref} for setup details. DPP shows outstanding performance; it also demonstrates the most stable results in terms of the variance between the runs. Arrows on the bottom indicate box plots being below the bottom boundary.}
  \label{fig:ll-B}
  \end{figure}

\begin{figure}[t!] 
    \centering
    \includegraphics[scale=.3]{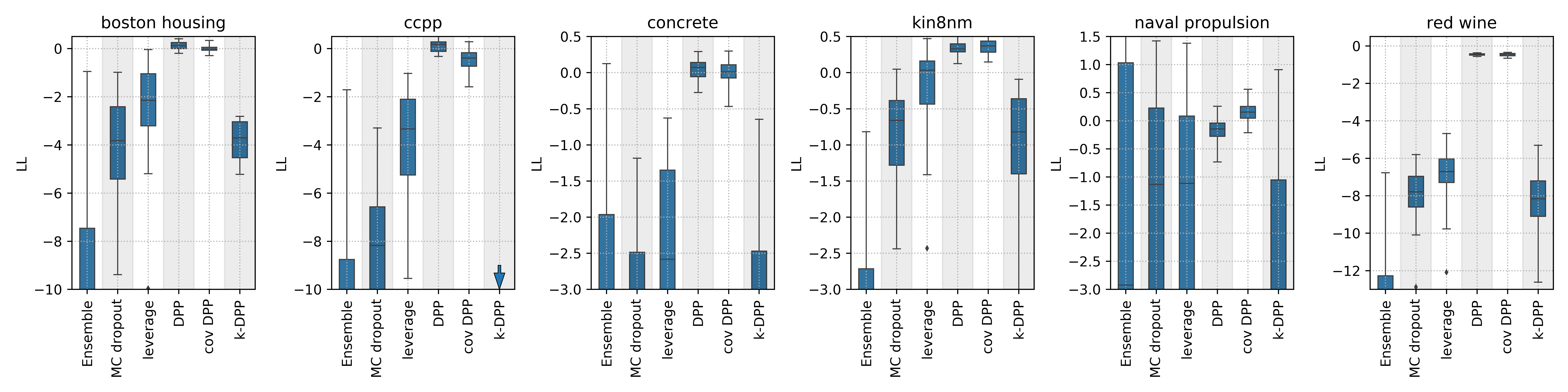}
    \caption{Log-likelihood across various UCI datasets for single models of NN UE for the large-NN experiment C with the reduced dropout rate, see Table \ref{tab:exp-ref} for setup details. Larger number of stochastic runs (30, 100) demonstrate the performance inferior to the 10 runs approach. DPP and k-DPP methods show stable dominance over other approaches. Arrows on the bottom indicate box plots being below the bottom boundary.}
  \label{fig:ll-C}
  \end{figure}

\begin{figure}[t!] 
    \centering
    \includegraphics[scale=.45]{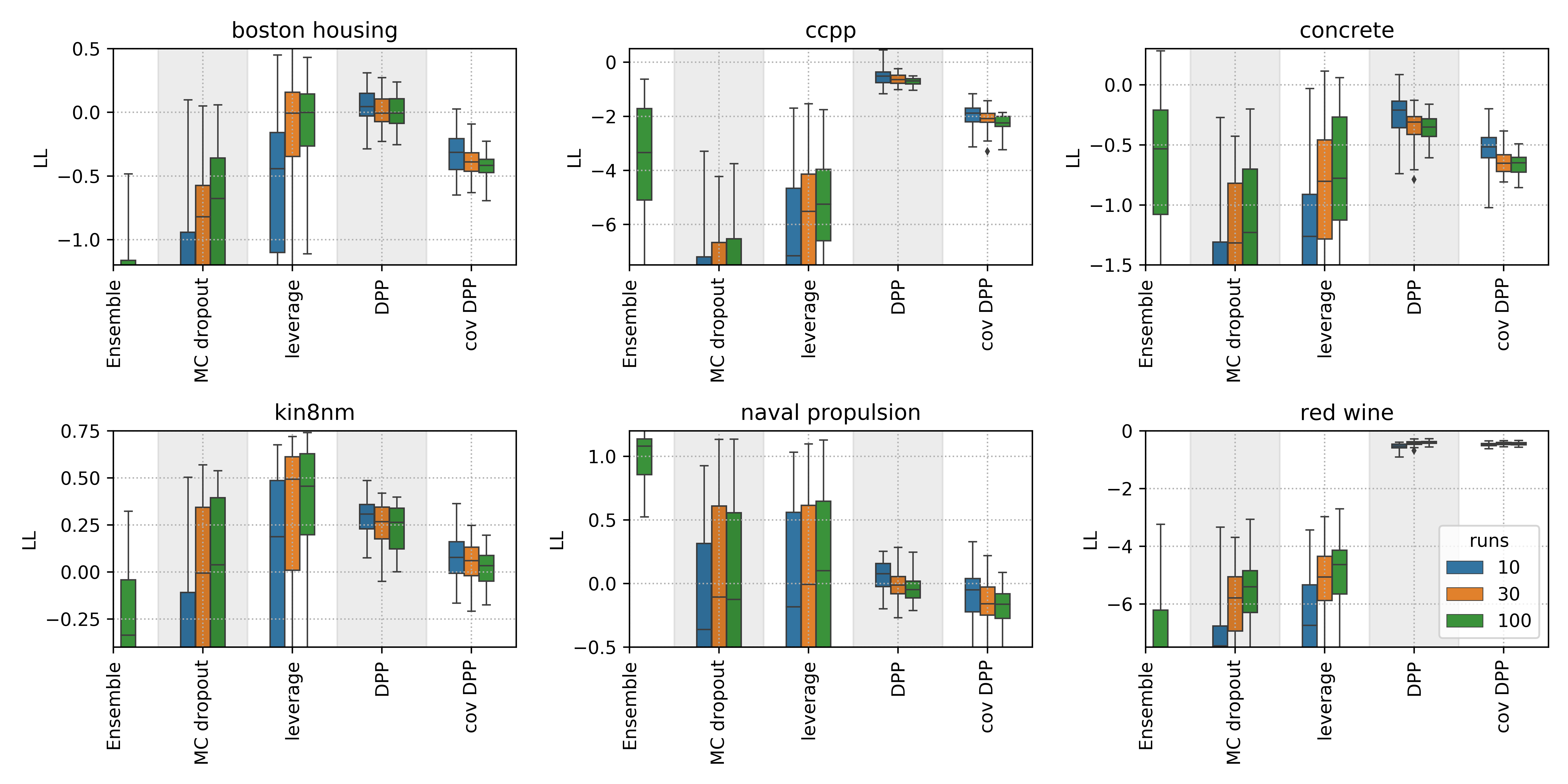}
    \caption{Log-likelihood across various UCI datasets for single models of NN UE for the large-NN experiment D, see Table \ref{tab:exp-ref} for setup details. DPP and leverage mask decorrelation shows the best results. Arrows on the bottom indicate box plots being below the bottom boundary.}
  \label{fig:ll-D}
  \end{figure}

  We have visualized the results for other experiments in Figures~\ref{fig:ll-B}, \ref{fig:ll-C} and~\ref{fig:ll-D}. 
  DPP-based methods show the best performance for majority of cases. For the very large (relatively to the datasets size) NN architecture, leverage score-based approach shows promising performance as well. 
  
%% file: version_2020/classification_suppl.tex

\subsection{Details on Models Training}
  In the classification tasks, neural networks were trained for 100 epochs maximum, with checking the error on the validation set every epoch: early stopping triggers if the error did not decrease for three consecutive checks (patience = 3). Batch size equals to 128, dropout applied after the hidden linear layer with a rate equal to 0.5. Cross entropy was used as a loss function, and optimization was performed with the standard setting of PyTorch Adam optimizer. For each dataset, we use different models. Below we denote convolutional layer with \(i\) input channels, \(j\) output channels as conv(i, j). Kernel size is 3x3 for each convolution.
  \begin{itemize}
    \item For the MNIST, we used simple convolutional network with layers conv(1, 16) - maxpool - conv(16, 32) - maxpool - linear(1152, 256) - dropout - linear(256, 10)
    
    \item For the CIFAR, we used VGG-alike network with layers conv(3, 16) - conv(16, 16) - maxpool(2, 2) - conv(16, 32) - conv(32, 32) - maxpool(2, 2) - conv(32, 64) - conv(64, 64) - maxpool(2, 2) - linear(1024, 128) - dropout - linear(128, 10)
    
    \item For the ImageNet, we used ResNet-18 with implementation and pretrained weights from PyTorch.
  \end{itemize}

  Ensembles of models for experiments below were trained separately on the same data from different weight initializations.

\subsection{Different Uncertainty Estimation Methods}
  For classification, one could use different uncertainty estimation measures. In the main part of the paper we provided results for BALD. The other popular uncertainty measures in the literature are
  \begin{itemize}
    \item \textbf{1 - maximum probability:} simple averaging  of probabilities over the ensemble and taking the one with maximum value:  \(s(\xv) = 1 - \max\limits_c ~ \bar{p}_T(y = c \mid \xv)\)
    \item \textbf{Variation ratio:} \(v(\xv) = 1 - \frac{f_{m}(\xv)}{T}\), where \(f_{m}(\xv)\) is the
    number of predictions for the most frequently chosen class by distinct runs of the model or members of the ensemble.
  \end{itemize}

\begin{figure*}[t!]
  \centering
  \begin{subfigure}[t]{0.23\textwidth}
    \centering
    \includegraphics[height=1.1in]{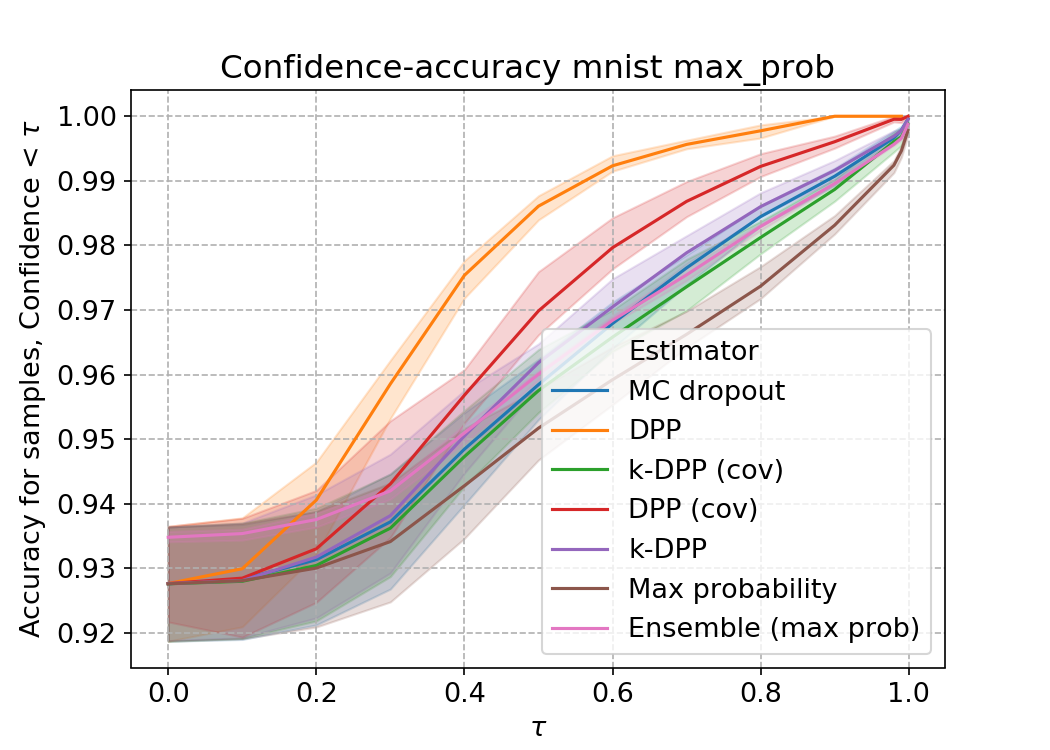}

  \end{subfigure}%
  ~ 
  \begin{subfigure}[t]{0.23\textwidth}
    \centering
    \includegraphics[height=1.1in]{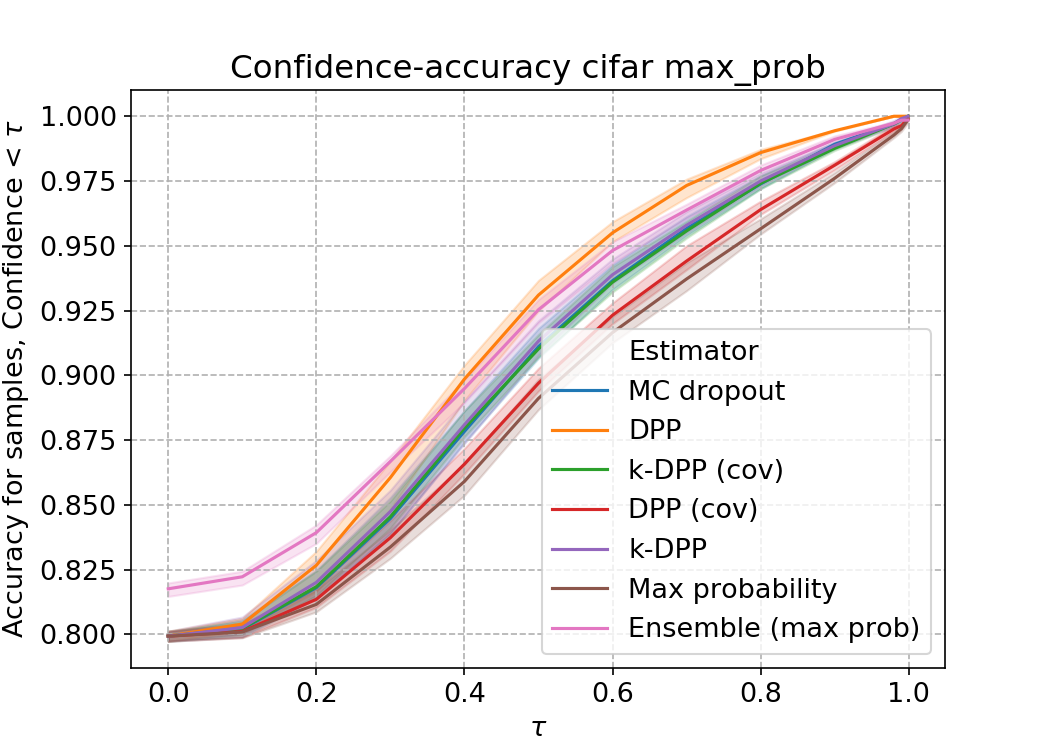}

  \end{subfigure}
  ~ 
  \begin{subfigure}[t]{0.23\textwidth}
    \centering
    \includegraphics[height=1.1in]{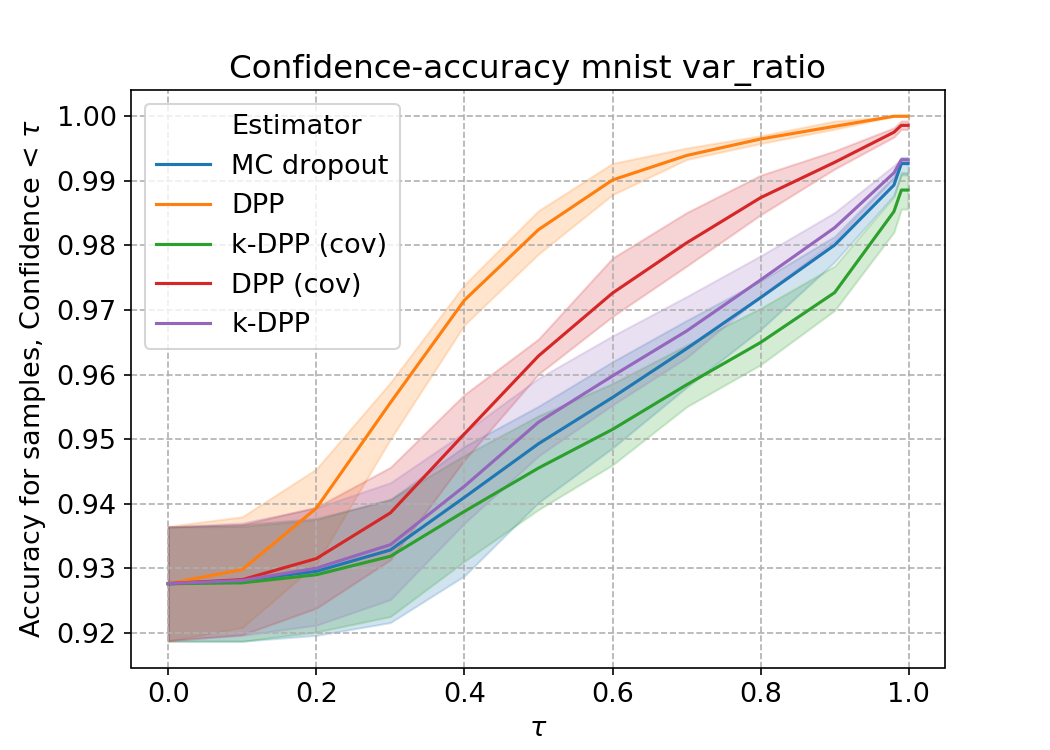}
  \end{subfigure}
  ~ 
  \begin{subfigure}[t]{0.23\textwidth}
    \centering
    \includegraphics[height=1.1in]{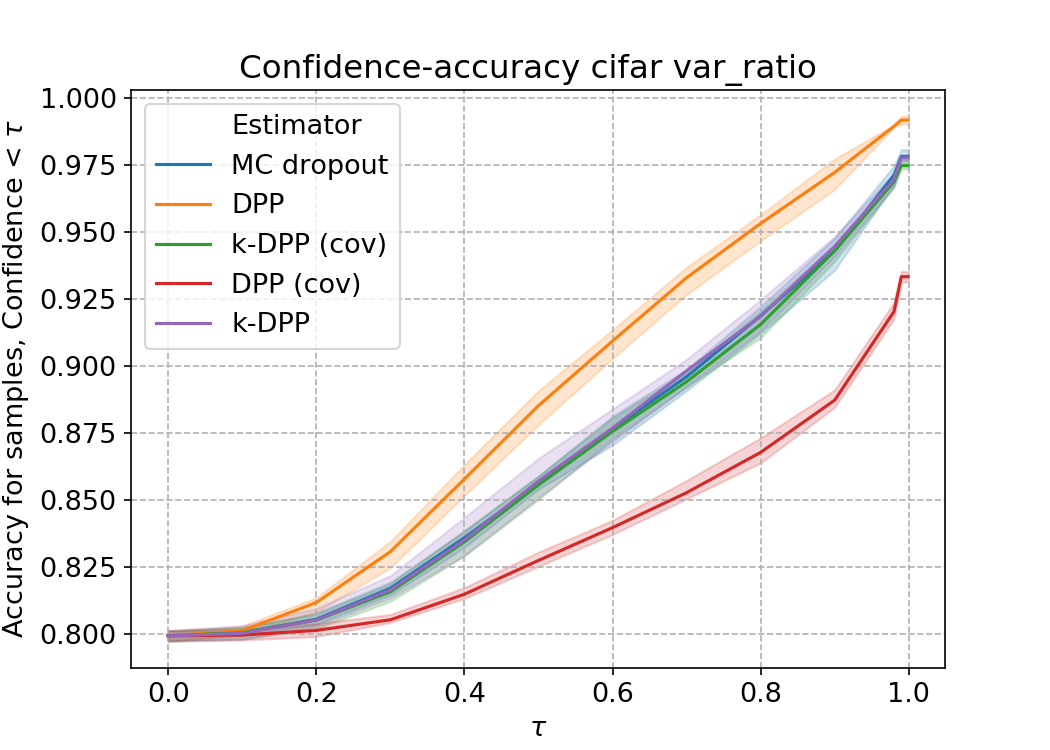}

  \end{subfigure}
  \caption{Accuracy-vs-confidence curve (the higher curve -- the better). We provide results for max probability and variation ratio measures for MNIST and CIFAR datasets.}
  \label{fig:classification-methods}
  \end{figure*}

\begin{figure*}[t!]
  \centering
  \begin{subfigure}[t]{0.23\textwidth}
    \centering
    \includegraphics[height=1.1in]{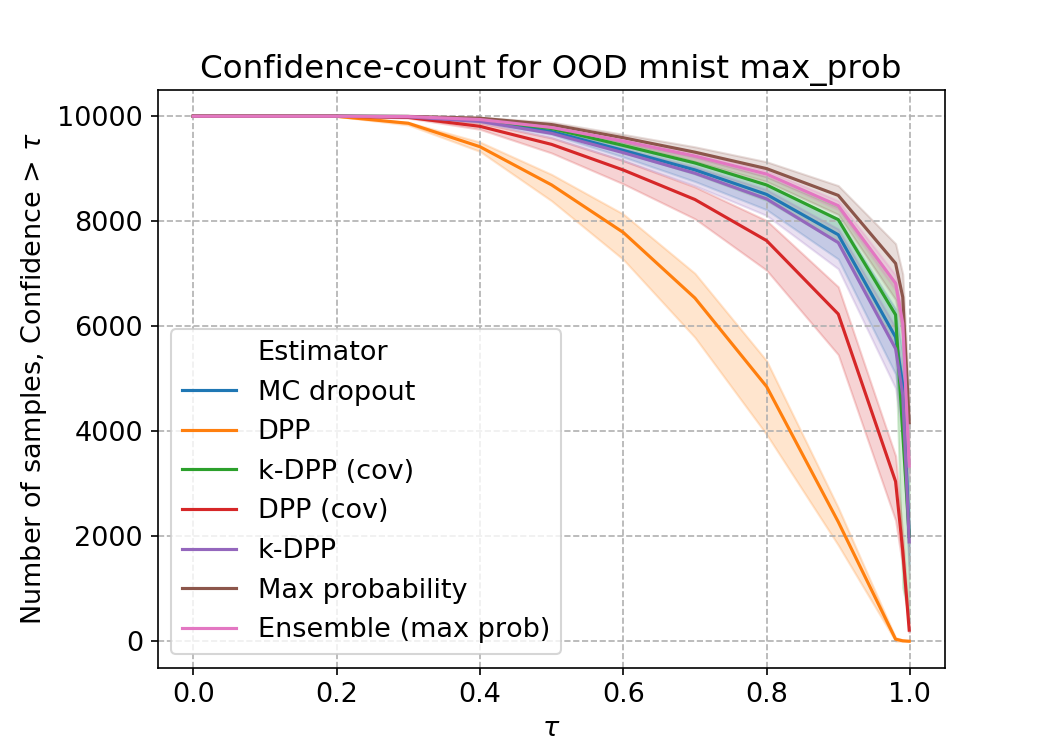}

  \end{subfigure}%
  ~ 
  \begin{subfigure}[t]{0.23\textwidth}
    \centering
    \includegraphics[height=1.1in]{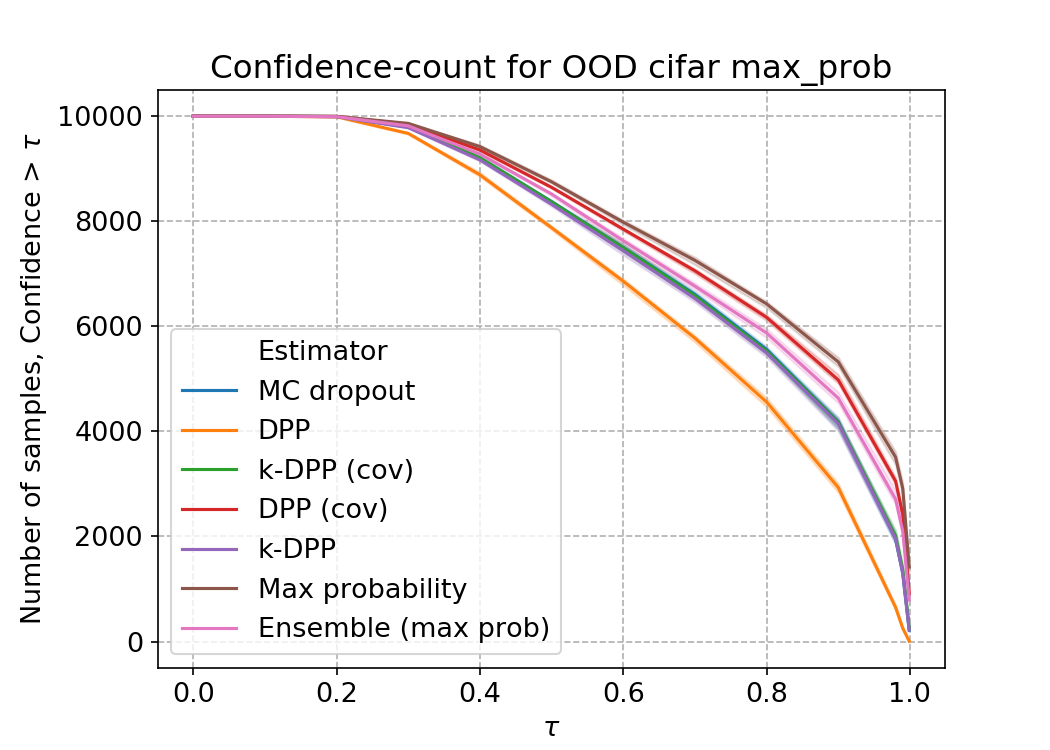}

  \end{subfigure}
  ~ 
  \begin{subfigure}[t]{0.23\textwidth}
    \centering
    \includegraphics[height=1.1in]{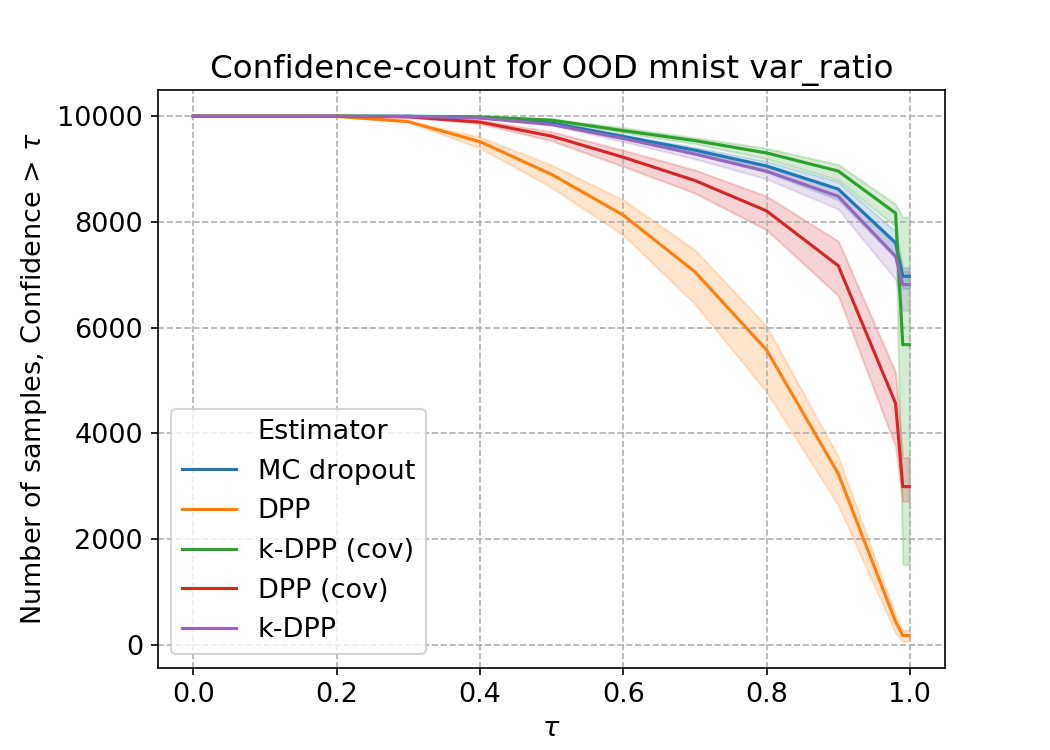}
  \end{subfigure}
  ~ 
  \begin{subfigure}[t]{0.23\textwidth}
    \centering
    \includegraphics[height=1.1in]{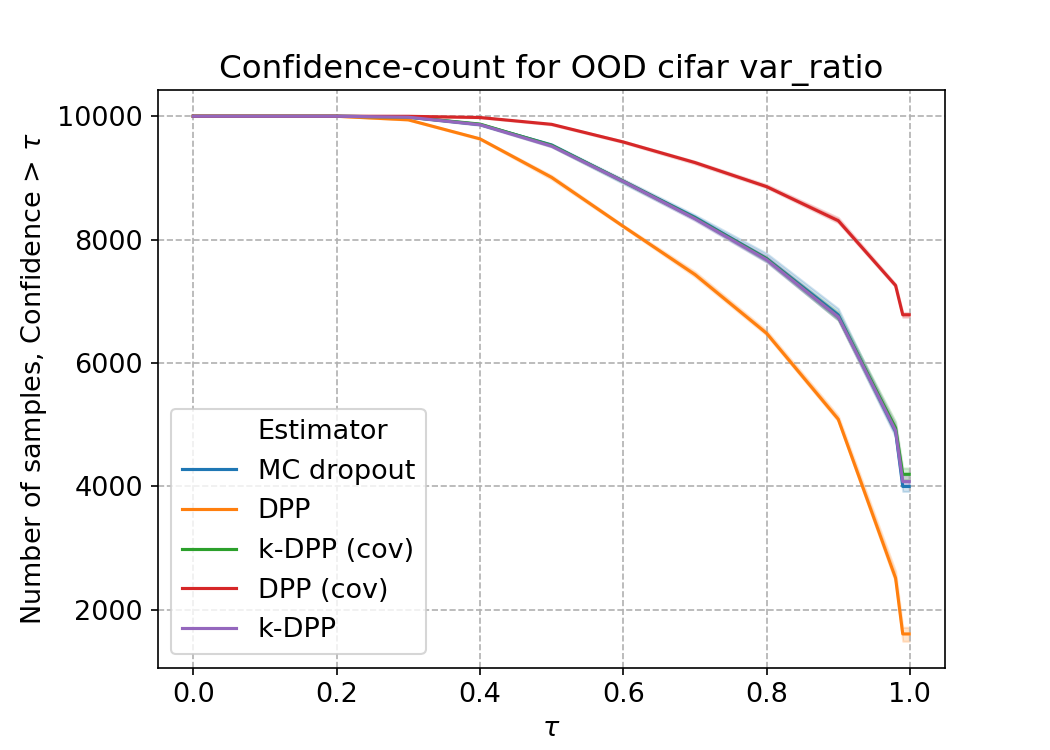}

  \end{subfigure}
  \caption{Count-vs-confidence curve for out-of-distribution data (the lower curve -- the better). We provide results for max probability and variation ratio measures for MNIST and CIFAR datasets.}
  \label{fig:classification-methods-ood}
  \end{figure*}

  In this section, we provide the results for these two methods (see Figure~\ref{fig:classification-methods}). For the probability, we average the predictions for all dropout masks. We also report the result for a single network without a dropout and for an ensemble of 20 independently trained networks. The same methods are applied for the OOD experiment (Figure ~\ref{fig:classification-methods-ood}). Note, here we provide the results for our methods both with correlation and covariance (cov) kernels. We can see that for these measures our DPP approach based on the correlation kernel outperforms the other approaches as well.

\subsection{ImageNet OOD Experiment}
\begin{figure}
    \centering
    \includegraphics[height=1.5in]{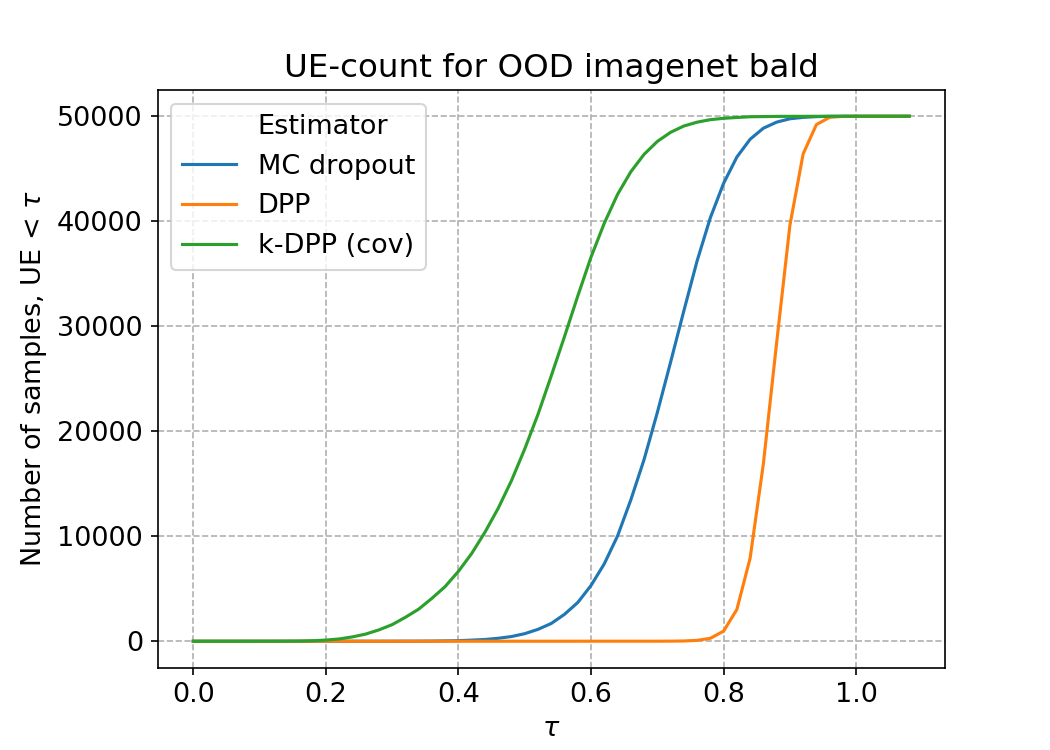}
    \caption{Count-vs-uncertainty curve for out-of-distribution data on ImageNet (the lower curve -- the better).}
    \label{fig:imagenet-ood}
  \end{figure}

  We present the out-of-distribution experiment for the ImageNet dataset here. As OOD images, we took the 50'000 samples from the CheXpert Small(chest radiography dataset)~\cite{irvin2019chexpert}. We used BALD uncertainty measure, the same as in the main part of the paper. The results are presented on Figure~\ref{fig:imagenet-ood}. The DPP method performs very well, while k-DPP is inferior to the MC Dropout.

\subsection{Scalability notes}
  One of our initial concerns was the scalability of the methods because DPP sampling complexity is up to \(N^3\), where \(N\) is the size of the layer. In practice, the overhead appears to be relatively small for real-world models, because the number of operations to compute dropout masks is negligible compared to the normal forward propagation in a large network. For the ImageNet experiment the difference was less than a few percent comparing to the Monte-Carlo dropout, see Table~\ref{tab:imagenet_performance}.
    
  \begin{table}[h!]
    \centering
    \caption{Time to calculate uncertainty with different methods on 5'000 sample images from ImageNet}
    \begin{tabular}{|c|c|c|c|c|c|c|}
      \hline
      & MC dropout & DPP & k-DPP \\
      \hline
      Inference time, s & 125.5$\pm$0.03  & 129.9$\pm$0.14  & 132.2$\pm$0.03 \\
      \hline
    \end{tabular}
  \label{tab:imagenet_performance}
  \end{table}